\documentclass[letterpaper, 10 pt, conference]{ieeeconf}  

\IEEEoverridecommandlockouts                             

\usepackage{array}
\usepackage{url}

\newcommand{\bigO}[1]{\mathcal{O}\left(#1\right)}

\newcommand{\model}{\textsc{GameOpt}\xspace}

\newcommand{\auctionname}{\textsc{Game}\xspace}
\newcommand{\optname}{\textsc{Opt}\xspace}
\newcommand{\main}{\textsc{GameOpt}\xspace}

\makeatletter
\newcommand\footnoteref[1]{\protected@xdef\@thefnmark{\ref{#1}}\@footnotemark}
\makeatother

\newcommand{\shorteq}{%
  \settowidth{\@tempdima}{-}
  \resizebox{\@tempdima}{\height}{=}%
}

\usepackage{amssymb,fge}

\usepackage[table]{xcolor}
\definecolor{sg}{HTML}{00ff7f}
\definecolor{lb}{HTML}{b0f5ef}
\definecolor{lg}{HTML}{9bfaa8}

\usepackage{amsmath, amssymb}
\usepackage{pifont}
\newcommand{\cmark}{\ding{51}}%
\newcommand{\xmark}{\ding{55}}%
\usepackage{subcaption}
\usepackage[utf8]{inputenc}
\usepackage[english]{babel}
\usepackage[table]{xcolor}
\usepackage[linesnumbered,ruled,vlined]{algorithm2e}
\usepackage[font=small,belowskip=-1.5pt]{caption} 
\usepackage{anyfontsize}

\usepackage{array}
\usepackage{graphicx}
\usepackage{amsfonts}
\usepackage[11pt]{moresize}
\usepackage{bm}
\usepackage{soul}
\usepackage{hhline}
\usepackage{multirow, makecell}
\usepackage{float}
\usepackage{booktabs,wrapfig}

\usepackage{amsthm}
\usepackage{color}
\usepackage{transparent}
\usepackage{url}
\usepackage{footmisc}
\usepackage{setspace}
\usepackage[colorlinks=true, citecolor=magenta]{hyperref}
\usepackage{mathtools}
\theoremstyle{plain}

\newtheorem{theorem}{Theorem}[section]

\begin{document}

\title{\main: Optimal Real-time Multi-Agent Planning and Control\\for Dynamic Intersections}

\author{Nilesh Suriyarachchi$^{1}$, Rohan Chandra$^{2}$, John S. Baras$^{1}$ and Dinesh Manocha$^{1,2}$
\thanks{$^{1}$Electrical and Computer Engineering Department, $^{2}$Computer Science Department, University of Maryland, College Park, Maryland, USA. Email: \tt\small\{nileshs,rchandr1,baras,dmanocha\}@umd.edu}
\thanks{\model page with videos: \scriptsize{\url{https://gamma.umd.edu/gameopt}}}
}



\maketitle

\begin{abstract}

We propose \model: a novel hybrid approach to cooperative intersection control for dynamic, multi-lane, unsignalized intersections. Safely navigating these complex and accident prone intersections requires simultaneous trajectory planning and negotiation among drivers. \model is a hybrid formulation that first uses an auction mechanism to generate a priority entrance sequence for every agent, followed by an optimization-based trajectory planner that computes velocity controls that satisfy the priority sequence. This coupling operates at real-time speeds of less than $10$ milliseconds in high density traffic of more than $10,000$ vehicles/hr, $100\times$ faster than other fully optimization-based methods, while providing guarantees in terms of fairness, safety, and efficiency. Tested on the SUMO simulator, our algorithm improves throughput by at least $25\%$, time taken to reach the goal by $75\%$, and fuel consumption by $33\%$ compared to auction-based approaches and signaled approaches using traffic-lights and stop signs.

\end{abstract}

\section{Introduction}
\label{sec: introduction}

Effectively navigating unsignalized intersections often requires carefully planning due to low visibility and complex maneuvers such as unprotected left turns. $40\%$ of all crashes, $50\%$ of serious collisions, and $20\%$ of fatalities occur at unsignalized intersections~\cite{grembek2018introducing}. The notion of introducing vehicle-to-infrastructure (V2I) communication capable connected autonomous vehicles (CAVs), to provide additional sensing and actuation points in the traffic flow, allows for the development of new algorithms to handle these scenarios. This has been applied successfully in complex traffic bottlenecks such as highway merging and traffic shock waves \cite{basic,Nilesh2021merge,Nilesh2021shockwave}. Recent research into leveraging cooperation among CAVs to achieve safe, fair, and efficient intersection control has provided promising results with solutions from many fields, including game theory, auctions, optimization, and deep learning. While each domain has its own advantages, no single solution has been shown to be able to achieve all the four key considerations of safety, efficiency, fairness, and real-time operation. Here, safety involves the prevention of collisions, efficiency involves maximizing the capacity of the intersection, and fairness involves treating all agents equitably. Real-time computation is also essential in this scenario, as the system must react to dynamic changes in this fast-paced environment. 

The general multi-agent intersection control problem can be broadly divided into two phases. The planning phase involves the selection of the optimal entrance sequence (order in which vehicles enter the intersection) and the control phase involves generating safe trajectories for all the vehicles in order to achieve the selected sequence. The system then transmits the output target trajectories to each vehicles' local controller for execution.
Here, sequence selection can become very difficult as the number of possibilities grows exponentially with the number of vehicles. This is further complicated in multi-lane scenarios, which should allow multiple non-conflicting vehicles to enter the intersection at the same time to achieve optimal throughput. Additionally, practical intersection scenarios are \textit{dynamic}, which means that the number of vehicles that request to cross the intersection changes with time.


\begin{figure} [t] 
\centering
\includegraphics[width=0.95\columnwidth]{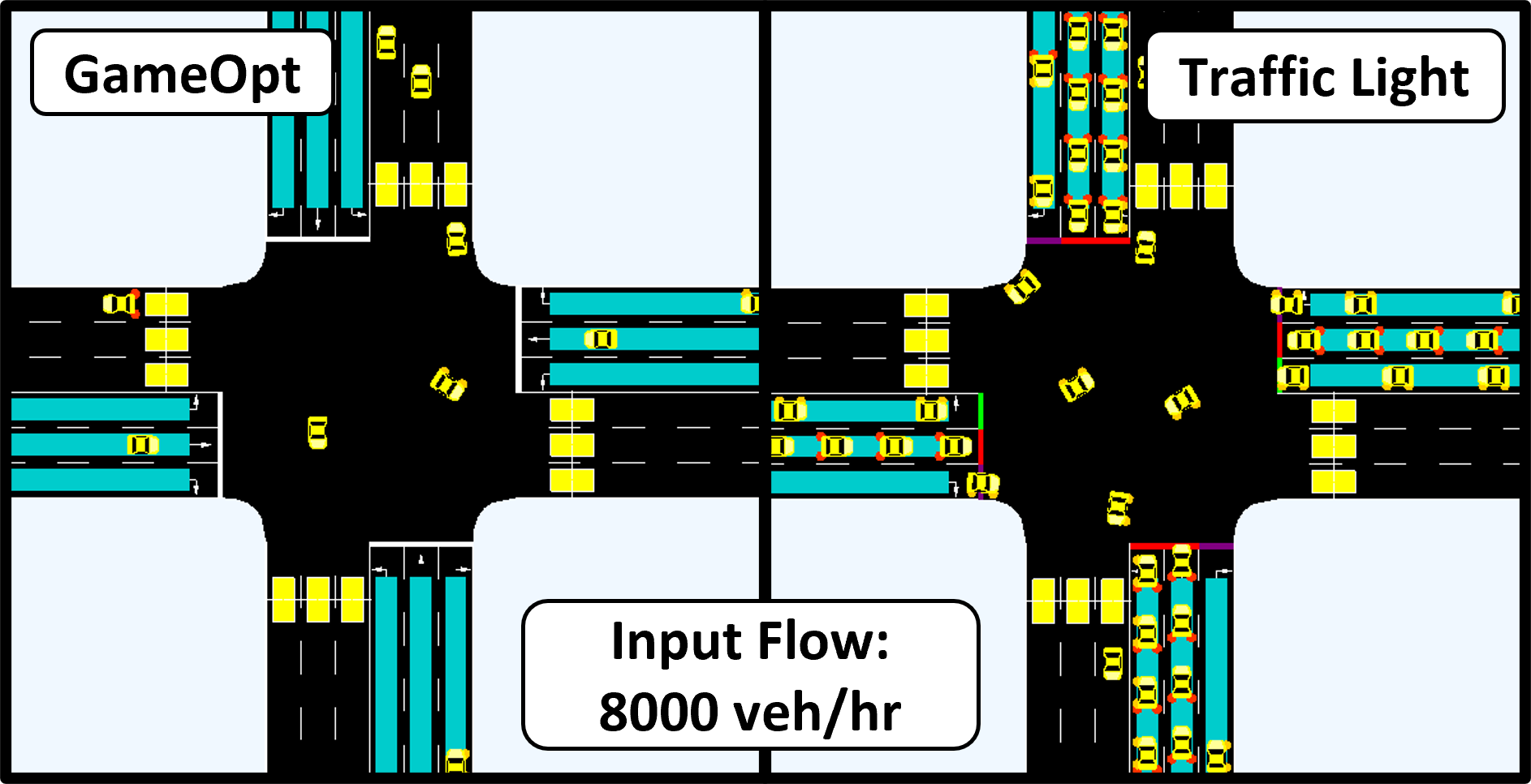}
\caption{\textbf{\model vs. Traffic Lights:} We present a new approach for optimal real-time planning and control in dynamic, multi-agent unsignalized intersections. In this figure, the light blue regions denote the control zone. We show that at identical input traffic flow levels, our approach outperforms even traffic light infrastructure, resulting in less queues and enabling smooth traffic flow.}
\label{fig: cover}
\vspace{-15pt}
\end{figure}

\vspace{-10pt}
\subsection{Related Work}
\label{subsec: related_work}
In Table~\ref{tab: comparison_table}, we compare our approach with the current state-of-the-art in navigating unsignalized intersection scenarios on the basis of optimality guarantees, dynamic intersection handling and real-world applicability through real-time computation capabilities. 

\subsubsection{Auctions}
Auction mechanisms have been used to navigate unsignalized intersections in real-time with fairness and feasibility guarantees~\cite{chandra2022gameplan, carlino2013auction}. But they work only in \textit{static} scenarios, where the number of participating agents is fixed, yielding poor efficiency and throughput in scenarios where the number of vehicles is variable. Incentive-compatible auctions such as Sayin et al.~\cite{sayin2018information} propose a mechanism in which agents are assigned turns based on their distance from the intersection and the number of passengers in the vehicle. Carlino et al.~\cite{carlino2013auction} and Rey et al.~\cite{rey2021online} propose a similar mechanism but use a monetary-based bidding strategy. Buckman et al.~\cite{gt6} integrate a driver behavior model~\cite{schwarting2019social} within the first-in first-out framework to incorporate human social preference. More recently, Chandra et al.~\cite{chandra2022gameplan} proposed an auction based on the driving behavior of the agents. 
 

\subsubsection{Game theory}
Game-theoretic approaches have also been used for autonomous vehicle navigation~\cite{li2020game, tian2020game}. Li et al.~\cite{li2020game} implement a stackelberg game in which one agent is the leader (modeled by first-out (FIFO) principle) and the other agent is a follower. Tian et al.~\cite{tian2020game}, on the other hand, use a recursive $k-$level approach in which strategies at current levels are derived from previous levels. However, all agents except the ego-agent at the first level are assumed as static.

\begin{table}[t]
    \centering
    \resizebox{.9\columnwidth}{!}{
    \begin{tabular}{ccccccc}
    \toprule[1.25pt ]
        Approach & Intersection & \multicolumn{3}{c}{Optimality} & Realtime\\
        \midrule
          &  &  Safety & Efficiency & Fairness &\\
          \cmidrule{ 3-5}
          Auctions& \cellcolor{red!25} Static & \cellcolor{red!25} \xmark & \cellcolor{red!25} \xmark & \cellcolor{green!25} \cmark &\cellcolor{green!25} \cmark\\
          Game Theory & \cellcolor{green!25} Dynamic & \cellcolor{red!25} \xmark & \cellcolor{red!25} \xmark & \cellcolor{green!25} \cmark &\cellcolor{red!25} \xmark\\
          Optimization & \cellcolor{green!25} Dynamic & \cellcolor{green!25} \cmark & \cellcolor{green!25} \cmark & \cellcolor{red!25} \xmark &\cellcolor{red!25} \xmark\\
          Deep Learning & \cellcolor{red!25} Static & \cellcolor{red!25} \xmark & \cellcolor{green!25} \cmark & \cellcolor{red!25} \xmark &\cellcolor{green!25} \cmark\\
          \midrule
        \textbf{This work} & \cellcolor{green!25} Dynamic & \cellcolor{green!25} \cmark & \cellcolor{green!25} \cmark & \cellcolor{green!25} \cmark &\cellcolor{green!25} \cmark \\

          \bottomrule[1.25pt ]
    \end{tabular}
        }
    \caption{\textbf{Comparison with prior work:} We highlight approaches for navigating unsignalized intersections based on the type of intersection, optimality guarantees, and real-time capability.}
    \label{tab: comparison_table}
    \vspace{-10pt}
\end{table}

\subsubsection{Optimization-based methods}

These methods compute optimal trajectories and provide safety guarantees, however, they are unable to perform in real-time due to high computational costs. Existing methods circumvent this issue by using unrealistic assumptions. Rios-Torres \textit{et al.} \cite{basic,Malik2019unconstrained} solve the optimization problem in real time using Hamiltonian analysis, which does not extend to the fully constrained problem. Computation is also restricted to a fewer vehicles in single lane roads, and no turning at intersections, to allow for close to real-time computing~\cite{Bian2020opt,Pei2021spaceopt,Seyed2018opt}. In Suriyarachchi \textit{et al.} \cite{Nilesh2021merge}, we applied a hybrid rule and optimization based control to the highway merging scenario to achieve real-time performance. However, the selection of sequences based on rules only provides a sub-optimal solution. 
Most of these methods do not scale well into multi-lane scenarios. 

\subsubsection{Deep learning}

Deep learning methods involving reinforcement learning~\cite{capasso2021end}
and recurrent neural networks~\cite{roh2020multimodal} learn a planning and control policy, to be used by agents approaching an intersection. In practice, these policies do not generalize well to different environments and often do not provide guarantees in terms of safety and fairness. Furthermore, most learning methods do not easily extend to dynamic multi-agent planning and control scenarios.

\subsection{Main Contributions:}


In the field of cooperative unsignalized multi-lane intersection control, existing work is unable to combine optimality guarantees in safety, efficiency, and fairness, with real-time performance. In order to bridge this gap, we develop a novel hybrid planning and control algorithm for navigating unsignalized dynamic intersections by combining auctions with optimal control. This is the first such framework, with implications that extend beyond intersection control to more general multi-agent systems, due to the challenging nature of dynamic mechanism design. Our main contributions are:
\begin{itemize}
\item \textit{Game-theoretic optimality:} The priority order sequence is fair, efficient, and tractable (Section~\ref{subsec: optimality}).
\item \textit{Safety guarantees:} The optimal trajectories satisfy the priority order with safety guarantees (Section~\ref{sec:optimization}).
\item \textit{Dynamic flow:} \model handles varying traffic flow in different arms of the intersection (Figure~\ref{fig: speed_inflow_analysis}).
\item \textit{Real-time computation:} \model operates on average at $1.16$ms with more than $10000$ vehicles per hour.
\item \textit{Multi-lane capability:} Increased performance by allowing many vehicles to enter intersection simultaneously.
\item \textit{Efficiency:} We outperform state-of-the-art by improving throughput, time-to goal, and fuel consumption on a realistic traffic simulator (Figure~\ref{fig: results}).

\end{itemize}

\section{Problem Formulation}
\label{sec: problem_formulation}

In this section, we formally define the fully automated unsignalized intersection planning and control problem as a large optimal control problem consisting of a hierarchy of two simpler decoupled optimization problems: mechanism design followed by vehicle trajectory planning. 
In a dynamic scenario consisting of $n$ vehicles, our goal is to compute velocities $v_i$ for each vehicle such that they autonomously navigate an unsignalized multi-lane intersection safely, fairly, and efficiently. We observe the following goals/constraints:

\begin{enumerate}
    \item \textit{Fairness}: The computed velocities, $v_i$, must maximize utility across all agents and be incentive compatible.
    \item \textit{Safety}: The trajectories generated by the optimizer must not result in collisions.
    \item \textit{Efficiency}: The overall algorithm must optimize throughput, time-to-goal, and fuel efficiency.
    \item The overall planning and control must not assume access to the objective or utility functions of other agents.
    \item The overall planning and control must not assume a dynamics model; rather, it should compute the dynamics for each agent assuming a simple forward motion model.
\end{enumerate}

The first step in this process involves defining the physical characteristics of an unsignalized intersection as well as the dynamics and control of the CAVs involved. 

\subsection{Modeling the Physical Intersection}
\label{subsec: physical_intersection}
\begin{figure} [h!] 
\centering
\includegraphics[width=.7\columnwidth]{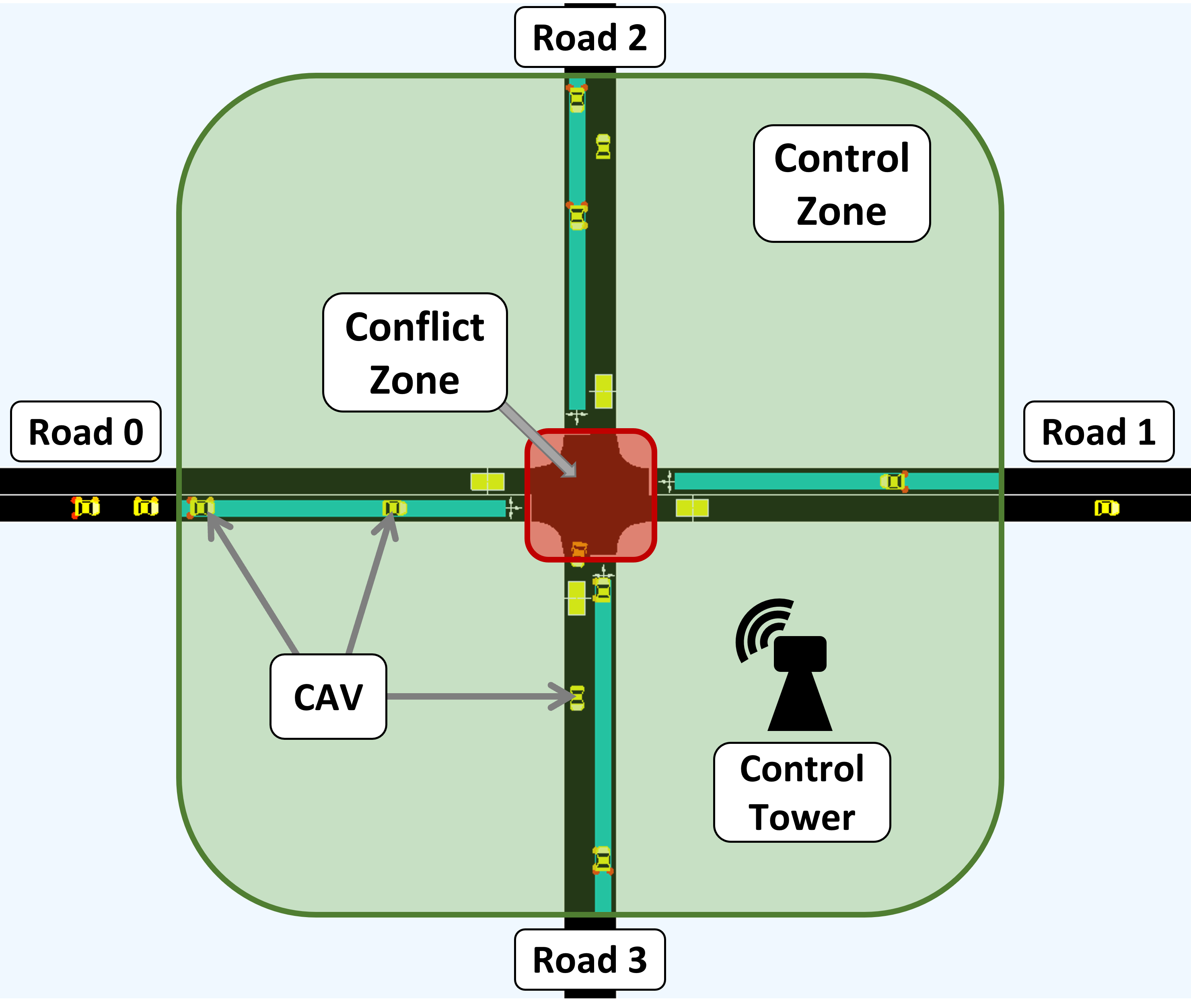}
\caption{\textbf{Regions of interest:} We highlight the control zone in which vehicles communicate with the control tower and the conflict zone in which vehicles cross over to their desired target road.}
\label{fig: zones}
\vspace{-7pt}
\end{figure}

We define an unsignalized intersection as a single-lane or multi-lane four-way crossing that is not regulated either by traffic signals or right-of-way rules. In Fig. \ref{fig: zones}, we show an example of a single-lane intersection used in our work. We define a control zone of length $L_c$ along each arm of the intersection; vehicles in the control zone can share state information and receive actuation commands using V2I communication protocols. In this research we assume zero transmission delay and that this V2I communication channel has perfect conditions. The center of the intersection is referred to as the conflict zone, where collisions are likely to occur. 
We denote the number of vehicles in the control zone in roads $0,1,2,3$ as $n_1,n_2,n_3,n_4$, respectively, with the total number of vehicles in the control zone being $n=n_1$+$n_2$+$n_3$+$n_4$.

Vehicles are autonomous and can freely choose between turning left, turning right, and going straight. The input flow rate of vehicles in each control zone of the intersection can be adjusted along with the ratio of vehicles that turn left, turn right, or go straight. 
We vary the input flow rate between $2,000$ vehicles per hour to $10,000$ vehicles per hour, where vehicles are generated according to a Poisson distribution.


\subsection{Vehicle Dynamics and Control Variables}
\label{sec: control_var_dynamics}
In real-world scenarios, vehicle dynamics are hard to model accurately due to their non-linearity. Standard practice in control theory suggests the use of a local controller, $z_i$, that produces low-level controls for the longitudinal and lateral motion of a vehicle according to the following equation, 

%
\begin{equation}
\label{eq:eq3}
    \frac{\delta s_i}{\delta t} = \mathcal{F}(t,s_i,z_i).
\end{equation}
\begin{figure*}[t] 
\centering
\includegraphics[width=.85\textwidth]{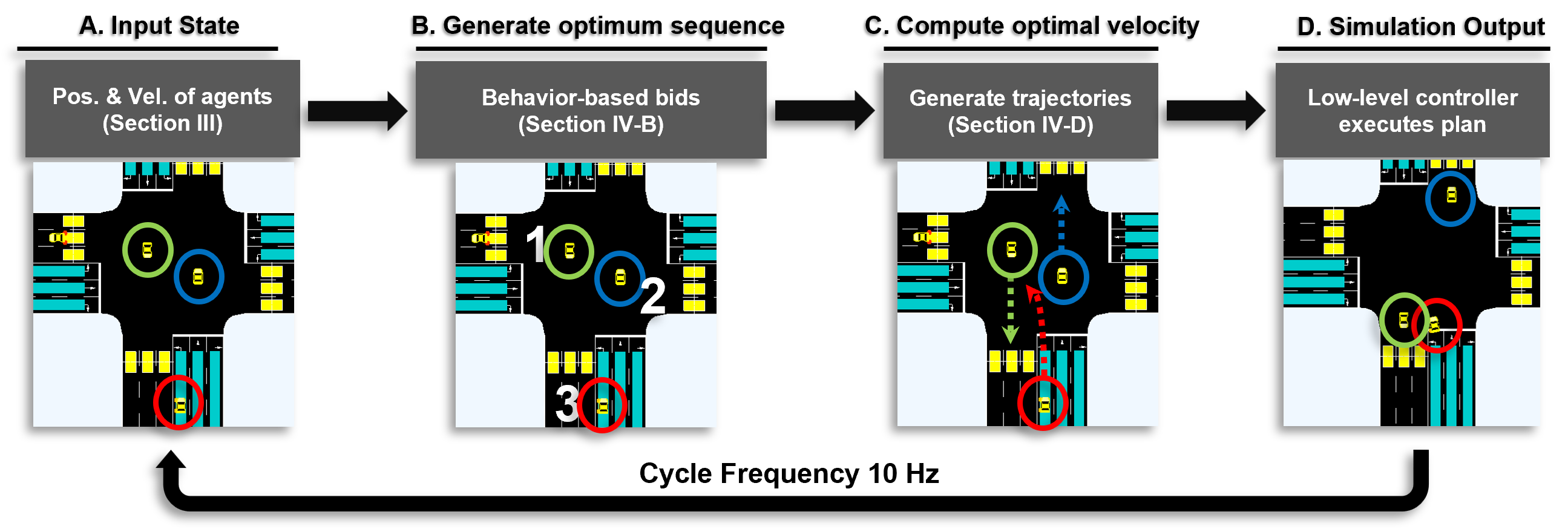}
\caption{\textbf{\model overview:} Our approach begins by reading in the positions and velocities of all agents in the control zone (blue). Our approach is hybrid; in the planning phase, \model collect the bids from every agent and generate an optimal priority sequence (Section~\ref{subsec: priority_order}). In the controls phase, we use an optimization-based trajectory planner to compute the optimal velocity for each agent that satisfies the priority order while simultaneously guaranteeing safety and real-time performance (Section~\ref{sec:optimization}).}
\label{fig: flow_diagram}
\vspace{-15pt}
\end{figure*}

\noindent Here, we represent $\mathcal{F}$ as the unknown non-linear dynamic function. $s_i$ denotes the distance of the agent $i$ from the intersection measured in the Frenet coordinate system. 

\noindent \textbf{Dynamics:} Since $\mathcal{F}$ is generally unknown, we define the high-level dynamics by the following velocity control scheme:
\begin{equation}
\begin{aligned}
\frac{\delta s_i}{\delta t}  &= v_i \\
v_i(t) &= \bm{u}_i(t) 
\end{aligned}
\label{eq:eq1}
\end{equation}
\noindent for $i\in\{1,\ldots,n\}$. 
 

\noindent \textbf{Control variables:} The control input, $\bm{u}_i(t)$, represents the command velocity for vehicle $i$. We compute the optimal value for $\bm{u}_i(t)$ with respect to safety, efficiency and reachability guarantees described in the following sections.

\noindent \textbf{State vector:} In addition to $s_i(t)$ and $v_i(t)$, we assign 
an indicator variable $l_c^i(t)\in\{0,1,\ldots,L_n-1\}$, which 
signifies which lane the vehicle is currently in, with $L_n$ representing the number of lanes on each road. We also assign a bidding value $b_i(t)$ to each vehicle according to \ref{subsec: priority_order}, which allows for the computation of the sequence order in which vehicles would cross the intersection. We also obtain the length $l_i$, and maximum acceleration $a^{max}_i$ and deceleration $a^{min}_i$ capabilities of each CAV $i$.
%
The CAVs are thereby completely defined by the state vectors: 
\begin{equation}
\label{eq:eq2}
    \bm{\Phi}_i(t) = [s_i(t), v_i(t), l_c^i(t), b_i(t), l_i, a^{max}_i, a^{min}_i]^\top 
\end{equation}

\noindent for $i\in\{1,\ldots,n\}$. We also define the speed limit, $\bar v$, and
a parameter tuple $M_s = \{M_{sr},M_{sl}\}$, which represent the safety margins needed to prevent rear-end and lateral collisions.

\subsection{Dynamic Traffic Intersection Planning \& Control Problem}
\label{sec: initial_formulation}
The dynamic traffic intersection problem involves computing suitable, collision-free, continuous trajectories for all vehicles approaching an intersection to achieve efficient and fair traffic flow. 
This problem can be formulated as an optimal control problem, with the objective of computing the optimal command velocities $\bm{u}_i$ for each CAV $i\in\{1,\ldots,n\}$, resulting in the minimization of the maximum time taken by each CAV to cross the intersection:

\begin{equation}
\begin{aligned}
&~~ \min_{\{\bm{u}_i\}} ~~~ \max_i ~ t_f^i \\
&s.t. \quad\quad \mathcal{C}(\{\bm{\Phi}_i\},\bar v, \mathbb{S}(M_s))
\end{aligned}
\label{eq:opt_time}
\end{equation}
\noindent where $t_f^i$ is the intersection crossing time of vehicle $i$. $\mathcal{C}$ is a set of constraints, discussed in Section~\ref{sec:optimization}, that depend on the state vectors $\{\bm{\Phi}_i\}$ of the CAVs, and safety requirements $\mathbb{S}(M_s)$ which prevent collisions between vehicles.

Equation~\ref{eq:opt_time} is, however, intractable to optimize directly because it implicitly assumes that agents are navigating the intersection according to the optimal priority order, $q^*$. Equation~\ref{eq:opt_time} can be equivalently solved by optimizing jointly over sequences, $q$, and the control velocities, $\bm{u}_i$, needed to implement the sequence.

\begin{equation}
\begin{aligned}
&~~ \min_{\{\bm{u}_i,q\}} ~~~ \sum_{i=1}^n 
    \lambda (\bm{u}_i-\bar v)^2 + (1-\lambda) (\bm{u}_i-v_i)^2 \\
&s.t. \quad\quad \mathcal{C}(\{\bm{\Phi}_i\},\bar v,\mathbb{S}(M_s),q)
\end{aligned}
\label{eq:opt_velocity}
\end{equation}
\noindent We solve this optimization problem in Section~\ref{sec:optimization}.\\

\noindent\textbf{Intractability of Equation~(\ref{eq:opt_velocity}):} The optimal control velocities, $\bm{u}_i$, depend on the sequence order $q$ of the vehicles entering the conflict zone. The naive solution is to solve a mixed-integer optimization problem for every possible entrance sequences $q$ by performing an exhaustive search. This process involves two sub-problems---generating a sequence $q$ followed by computing the optimal velocity commands $\bm{u}_i$ for that sequence. The total number of possible sequences $q$ grows exponentially with the number of vehicles in the control zone which would take the form $\bigO{\frac{(n_1+n_2+n_3+n_4)!}{n_1!n_2!n_3!n_4!}}$, where $n$=$n_1$+$n_2$+$n_3$+$n_4$ is the number of vehicles in the control zone. Therefore, current methods to solve Equation~\ref{eq:opt_velocity} are intractable.

We propose the use of auctions to obtain an optimal priority sequence order $q^{*}$ in $\bigO{n\log n}$ time, significantly reducing the combinatorial runtime complexity of the mixed-integer formulation. We run the optimization over $\bm{u}_i$ for the selected optimal sequence $q^{*}$ which can be solved in real time.

\subsection{Auction Theory}
\label{subsec: SSA_background}

Auctions are a central theme in mechanism design. An auction allocates $K$ items among $n$ agents. Each agent $i$ has a private valuation $\bm{\zeta}_i$ and submits a bid $\bm{b}_i$ to receive at most one item of value $\bm{\alpha}_i$. In any auction, we have $\bm{b}_1 > \bm{b}_2 > \ldots > \bm{b}_K$ and $\bm{\alpha}_1 > \bm{\alpha}_2 > \ldots > \bm{\alpha}_K$. Each auction has an allocation rule (who gets what), a payment rule (who pays what), and a utility function (how much value an agent gets based on their bid).   
In a sponsored search auction, which is the model we are using in this work, the allocation rule is that the agent with the $j^\textrm{th}$ highest bid is allocated the $j^\textrm{th}$ most valuable item, $\bm{\alpha}_i$. The quasi-linear utility $\bm{\theta}_i$~\cite{roughgarden2016twenty} incurred by $i$ is as follows,

\begin{equation}
 \bm{\theta}_i (\bm{b}_i) =  \bm{\zeta}_i \bm{\alpha}_i - \sum_{j=i}^k \bm{b}_{j+1} \left( \bm{\alpha}_j - \bm{\alpha}_{j+1} \right).
 \label{eq: utility_template}
\end{equation}

\noindent The term $\bm{\zeta}_i \bm{\alpha}_i$ denotes the time reward gained by $i$ by moving on the $i^\textrm{th}$ turn. The payment term, $\sum_{j=i}^k \bm{b}_{j+1} \left( \bm{\alpha}_j - \bm{\alpha}_{j+1} \right)$, represents the risk~\cite{wang2020game} associated with moving on that turn. It follows that an allocation of a conservative agent to a later turn (smaller $\bm{\alpha}$) also presents the lowest risk and vice-versa.



\section{\model: Methodology}
\label{sec: gameopt}

We present our overall approach in Figure~\ref{fig: flow_diagram}. Computation begins by receiving the positions, velocities and initial bids from all agents in the control zone via V2I communication (Section~\ref{sec: problem_formulation}). In the following planning phase, this information along with the processed bids are used to generate an optimal priority order or sequence, which determines the order in which agents will enter the intersection (Section~\ref{subsec: priority_order}). The control phase of our approach then involves using an optimization-based trajectory planner to compute the optimal velocity (Section~\ref{sec:optimization}) for each agent, which satisfies the priority order while simultaneously guaranteeing safety and real-time performance.

\subsection{\auctionname: Selecting the Priority Order, $q^*$}
\label{subsec: priority_order}

Choosing an optimal ordering for agents to navigate unsignalized traffic scenarios is equivalent to allocating each agent a turn in which they would cross the intersection. Such an allocation depends on the incentives of the agents which, in many cases, are not known apriori. Auctions model the incentives of agents in unsignalized traffic scenarios using an optimal combination of bidding, allocation, and payment strategies. In the rest of this section, we present the auction framework for generating an optimal priority order, followed by an analysis of its fairness.

The SSA (Section~\ref{subsec: SSA_background}) is an ideal mechanism to generate a priority order for the agents.\textit{ The agent with the highest priority bid is allowed to navigate the scenario first, followed by the agent with the next highest priority, and so on.} This algorithm runs in polynomial time, since the main computation at this stage is dominated by sorting the agents' bids~\cite{clrs}. The priority value of an agent is its true value, $\bm{\zeta}_i$, and our approach accommodates many different metrics that can be used for computing the priority value for an agent. For instance, our approach can work with both driver behavior-based~\cite{chandra2022gameplan}, distance-based~\cite{gt6}, and monetary-based~\cite{carlino2013auction} bidding strategies. We compute the priority value based on its velocity and its distance from the intersection. More specifically, agents that are closer to the intersection and have higher velocities have higher priority to cross the intersection. 

More formally, recall that $s_i(t)$ measures the distance of agent $i$ from the intersection, and we use $\tau_i$ to denote the time in which $i$ shall reach the intersection, based on its current velocity and $s_i(t)$. Then, the priority value for agent $i$ is,

\begin{equation}
    \bm{\zeta}_i = s_i(t) \times (c - \tau_i),
    \label{eq: priority_bid}
\end{equation} 

\noindent where $c$ is a constant. To avoid a single lane dominating the intersection (due to a constant flow of fast moving vehicles) and to prevent congestion, we reward agents that have been waiting in queue for a long time. Thus, $\bm{\zeta}_i \gets w_i\bm{\zeta}_i$, where $w_i$ represents the waiting time reward value. Such a strategy of mitigating congestion has been commonly used in auction-based approaches~\cite{carlino2013auction}. Finally, to deal with multi-agent dynamic traffic, we implement an overflow strategy where, if an agent with a higher priority is behind an agent with a lower priority, then the former transfers some amount of their bid to the latter. This phenomenon is commonly observed in agent-based motion models such as the Social Forces model.

In~\cite{chandra2022gameplan}, the authors show that the sponsored search auction is game-theoretically optimal for static intersections consisting of $1$ vehicle on each arm of the intersection. In this work, we extend the proof for dynamic traffic intersections where we must take into account three factors--$(i)$ multiple vehicles on each arm of the intersection and $(ii)$ lower priority vehicles blocking higher priority vehicles, and $(iii)$ providing waiting time rewards to agents.

\subsubsection*{\ul{Fairness Analysis of $q^*$}}
\label{subsec: optimality}

From a game-theoretic perspective, fairness implies that no agent is incentivized to ``cheat'' or, in other words, this occurs when the dominant strategy for each agent is to bid their true valuation $\bm{\zeta}_i$. Incentivizing traffic-agents to bid their true value ($\bm{b}_i = \bm{\zeta}_i$) as a dominant strategy is known as incentive compatibility~\cite{roughgarden2016twenty}. Chandra et al.~\cite{chandra2022gameplan} showed that SSAs implemented at static intersections with a single agent on each arm are incentive-compatible. We extend the theoretical framework of static SSAs to dynamic intersections with multiple vehicles, taking into account \textit{overflow}, which is when a lower priority vehicle blocks a higher priority one.

\begin{theorem}
\textbf{Incentive compatibility for dynamic intersections:} For each agent $i$ for $i=1,2,\ldots,n$ and $n = n_1 + n_2 + n_3 + n_4$ where $n_j$ represents the number of vehicles on the $j^\textrm{th}$ arm, bidding $\bm{b}_i = \bm{\zeta}_i$ is the dominant strategy. 
\label{thm: incentive_compatibility}
\end{theorem}

The next desired property in a fair auction is welfare maximization~\cite{roughgarden2016twenty} which maximizes the total time reward (Section~\ref{subsec: SSA_background}) earned by every agent. 
We show that SSAs maximize social welfare for dynamic intersections as well.

\begin{theorem}
\textbf{Welfare maximization for dynamic intersections:} Social welfare of an auction is defined as $\sum_i \bm{\zeta}_i\bm{\alpha}_i$ for each agent $i$ for $i=1,2,\ldots,n$ and $n = n_1 + n_2 + n_3 + n_4$ where $n_j$ represents the number of vehicles on the $j^\textrm{th}$ arm. Bidding $\bm{b}_i = \bm{\zeta}_i$ maximizes social welfare for every agent. 
\label{thm: Welfare_Maximizing}
\end{theorem}

Finally, we propose a novel strategy to address overflow by transferring a portion of the higher priority agent's bid to the lower priority agent i. We denote such a transfer by $(\hat a, \hat b = a \xrightarrow{c} b)$, where $\hat a = a-c, \hat b = b+c$. More formally, 

\begin{theorem}
\textbf{Overflow prevention:} In a current SSA, suppose there exists $ (i,j) \in [n_k] \times [n_k], k=1,2,3,4$, such that $p_i > p_j$ and $s_i[t] > s_j[t]$. Let $ \bm{\hat \zeta}_i, \bm{\hat \zeta}_j = (\bm{\zeta}_i \xrightarrow{q} \bm{\zeta}_j)$. If 
\[ q < \bm{\zeta}_i \left( 1-\frac{\alpha_{i+m}}{\alpha_i}\right) - \sum_{s=i}^{i+m-1}\bm{\zeta}_{s+1}\left( \frac{\alpha_s - \alpha_{s+1}}{\alpha_i} \right),\] 

\noindent the new SSA with $\bm{\hat \zeta}_i, \bm{\hat \zeta}_j$ as the new priority values for $i,j$ is incentive compatible.
\label{thm: overflow}
\end{theorem}

As a final remark, Carlino et al.~\cite{carlino2013auction} show that multiplying a bid by a waiting time reward does not change the incentive compatibility of the auction. We defer the proofs of Theorems~\ref{thm: incentive_compatibility},~\ref{thm: Welfare_Maximizing}, and~\ref{thm: overflow} to \cite{gameoptPage}.

\subsection{Multi-lane Intersection Planning and Control }
\label{subsec: conflict_resolution}

\begin{figure}[t]
\begin{subfigure}[h]{.9\columnwidth}
\centering
    \includegraphics[width = 0.9\textwidth]{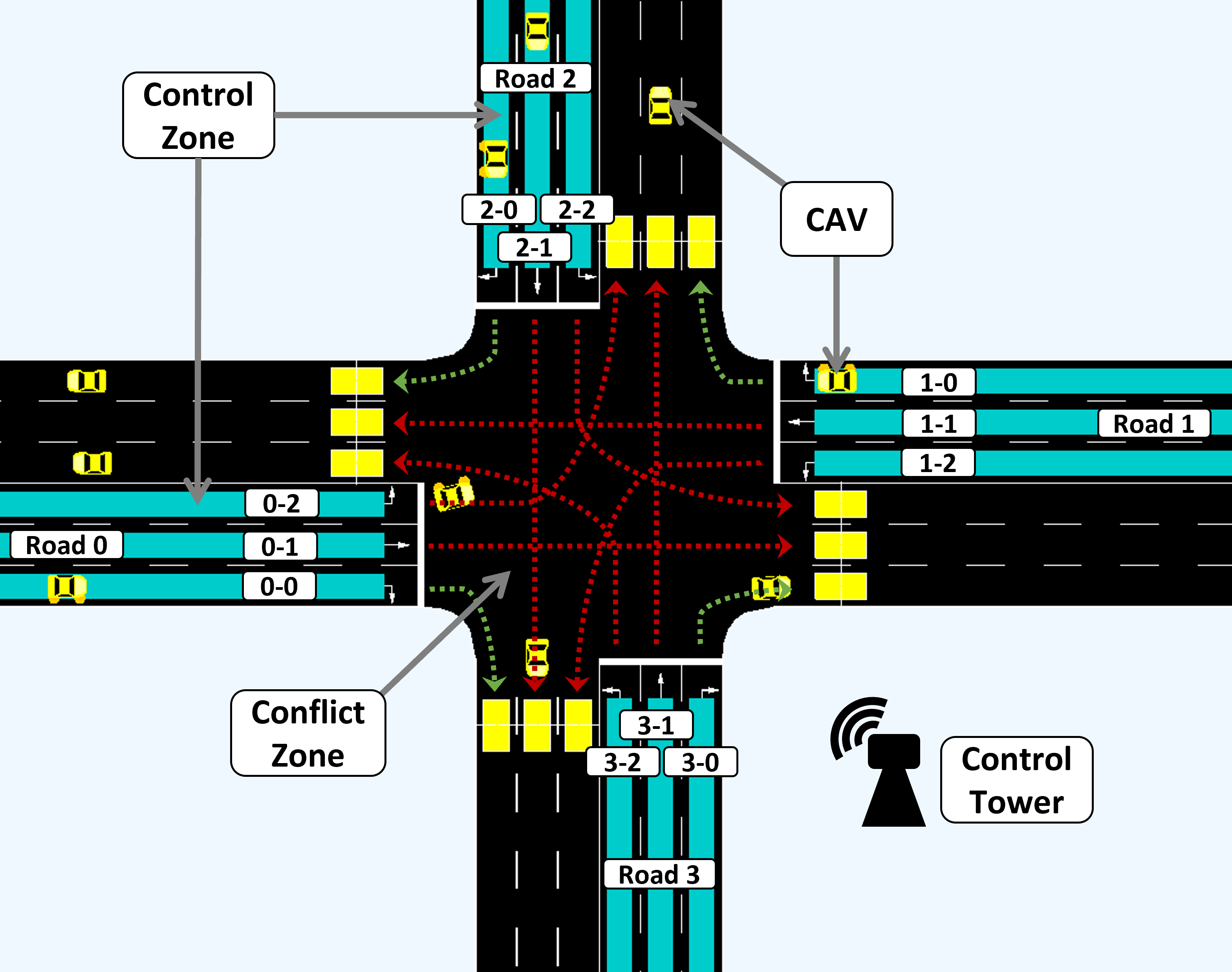}
        \caption{Multi-lane intersection structure with lane-based groups}
        \label{fig: conflict_groups}
\end{subfigure}
  \begin{subfigure}[h]{\columnwidth}
  \vspace{5pt}
    \centering
\resizebox{.6\columnwidth}{!}{
\begin{tabular}{cc}
\toprule[1.5pt]
{\textbf{Lane group}} & {\textbf{No-conflict group set}} \\ 
\midrule
0-1 & 0-2, 1-1, 2-2 \\
0-2 & 0-1, 1-2, 3-1 \\
1-1 & 0-1, 1-2, 3-2 \\
1-2 & 0-2, 1-1, 2-1 \\
2-1 & 1-2, 2-2, 3-1 \\
2-2 & 0-1, 2-1, 3-2 \\
3-1 & 0-2, 2-1, 3-2 \\
3-2 & 1-1, 2-2, 3-1 \\
\bottomrule[1.5pt]
\end{tabular}
}
    \caption{Non-conflicting trajectory groups at intersection}
    \label{table: grouping}
    \end{subfigure}
\label{fig: conflict_figure}
\caption{\textbf{Conflict handling at multi-lane intersections}. We propose a novel strategy for conflict handling using vehicle grouping based on non-colliding trajectories.}
\vspace{-15pt}
  \end{figure}

A key capability of our method is its ability to handle traffic effectively in multi-lane intersections. The extension from single lane to multi-lane is not straight forward. Many assumptions that can be made in single lanes (such as one vehicle at a time in the conflict zone) do not hold up in a multi-lane scenario. Therefore, we introduce a system to classify all vehicles in the control zone into groups based on their desired trajectory (origin road and intention) in order to handle conflicts among vehicles accurately. Group \textit{x-y: x=road\_num, y=intention} (Here intention values are: \textit{0=right\_turn, 1=go\_straight, 2=left\_turn}). For example, a vehicle from road 1 taking a left turn would be in group 1-2. Fig. \ref{fig: conflict_groups} shows the trajectories that vehicles in each lane can take, along with the labeling used to allocate groups based on these trajectories, marked on each of the lanes.

Here, we notice that vehicles taking right turns at the intersection (groups 0-0, 1-0, 2-0, 3-0) do not conflict with any other vehicle groups. Therefore, they are allowed to enter into the intersection freely, and the constraints in the optimization in \ref{sec:optimization}, reflect this. For each of the remaining lane groups, we need to identify which other lane groups can be allowed to enter the intersection at the same time. We denote this collection as the non-conflicting group set for each corresponding main lane group, as shown in Table \ref{table: grouping}. 
For example, consider the main lane group 0-1. The groups with trajectories that do not conflict with this main group are  0-2, 1-1, and 2-2 along with all the right turn groups (0-0, 1-0, 2-0, 3-0).
What this effectively means is that any vehicles belonging to the non-conflicting group set of a main lane group will be allowed to enter the intersection at the same time along with a vehicle of the main lane group. Note that vehicles in the right turn at intersection groups (0-0, 1-0, 2-0, 3-0) are directly marked as non-conflicting and are included in the non-conflicting groups list for all vehicle groups by default.

\subsection{\optname: Computing Optimal Velocities} 
\label{sec:optimization}

\noindent\textbf{The Optimization Problem:}

Based on the optimal priority sequence $q^*$ from Section~\ref{subsec: priority_order}, the optimal command velocities, $\bm{u}_i$, for each vehicle are computed by optimizing Equation~\ref{eq:opt_velocity} for a fixed sequence. We restate the objective function here for convenience:
\begin{equation}
\mathcal{J}(\bm{u}_i|q^*) = \min_{\{\bm{u}_i\}} \sum_{i=1}^n \lambda (\bm{u}_i-\bar v)^2 + (1-\lambda) (\bm{u}_i-v_i)^2
\label{eq:J}
\end{equation}

\noindent The first quadratic term of Equation~\ref{eq:J} prevents $\bm{u}_i$ from deviating from the speed limit  $\bar v$ (results in improved throughput). The second quadratic term minimizes the control effort applied to optimize fuel efficiency. Below we discuss constraints on safety and the influence of $q^*$ on Equation~(\ref{eq:J}).\\

\noindent \textbf{Constraints on Safety and Compliance with $\mathbf{q^*}$:} 

We consider both longitudinal (rear-end) collisions in the control zone and collisions in the conflict zone. Rear-end collisions along each of the lanes in the control zone are prevented by ensuring all vehicles in the same lane maintain a safe distance between each other. Formally,

\begin{equation} 
\Big | s_j^k(t+1)-s_{j\prime}^k(t+1)\Big | \geq l_{j} + M_{sr}
\label{eq: longitudinal_collision}
\end{equation}

\noindent for all $k$ lanes in the control zone and for all agents $j,j\prime$ with $j\neq j\prime$ in the $k^\textrm{th}$ lane. $l_j$ represents the length of the agent $j$ and $M_{sr}$ is a safety margin added to maintain a safe gap between vehicles and to account for imperfections in sensing and actuation. We compute the future position $s_i(t+1)$ using,

\begin{equation}
\label{eq:eq_cons_3_1}
    s_i(t+1) = s_i(t) - \Delta t \left (\frac{v_i(t)+\bm{u}_i(t)}{2}\right)
\end{equation}

\noindent Here, $\Delta t$ is defined as the planning time step of the controller.

Next, we also need to ensure that no collisions occur inside the conflict zone. This task reduces to ensuring no vehicles belonging to conflicting lane groups (\ref{subsec: conflict_resolution}) enter the intersection simultaneously. Therefore, for each vehicle $i$ in the sequence $q^*$, paired with a vehicle $j$ in a conflicting group where $i$ has higher priority over $j$, we add the constraint,

\begin{equation}
        t_c^i + \frac{(l_i + M_{sl})}{\bm{u}_i} \leq t_c^{j}
    \label{eq: lateral_collision}
\end{equation}

\noindent where $t_c^i= \frac{s_i(t+1)}{\bm{u}_i}$ represents the time at which $i$ would arrive at the intersection. The above constraint ensures that command velocities $\bm{u}_i$ are chosen such that vehicles proceed through the conflict zone in the order of the sequence under consideration $q^*$. Note that this constraint is applied to every conflicting pair of vehicles; non-conflicting vehicles are allowed to enter the intersection at the same time, increasing the efficiency of the system. The safety margin $M_{sl}$ corresponds to the width of the intersection and prevents lateral collisions occurring when conflicting vehicles cross the intersection. 

The safety constraints with respect to ordering $q^*$ in Equations~(\ref{eq: longitudinal_collision}) and~(\ref{eq: lateral_collision}) result in an intractable mixed-integer quadratic programming (MIQP) optimization problem. This problem is further compounded due to lane indexing, which introduces a new variable indicating the agent's lane. We remove the dependency over the lane index by separating the constraints for each lane and using the multi-lane conflict resolution scheme introduced in~\ref{subsec: conflict_resolution}. This results in simplification of Equations~(\ref{eq: longitudinal_collision}) and~(\ref{eq: lateral_collision}), reformulating the original MIQP as a quadratic programming (QP) optimization problem; this QP can be solved in real-time. More formally, Equation~(\ref{eq: longitudinal_collision}) can be simplified as,

\begin{equation}
\label{eq: long_linear}
    \begin{aligned}
        \bm{u}_{j}^k-\bm{u}_{j+1}^k \geq & \left(v_{j+1}^k-v_{j}^k\right) \\
        & + \frac{2}{\Delta t}\left(s_{j}^k-s_{j+1}^k+l_{j}+M_{sr}\right) 
    \end{aligned}
\end{equation}

\noindent for all $k$ lanes in the control zone and for all $j,j+1$ consecutive vehicles belonging to lane $k$. And Equation~(\ref{eq: lateral_collision}) can be simplified as,

\begin{equation}
\label{eq: lateral_linear}
        \bm{u}_{j}\left(s_i-\frac{\Delta t}{2}.v_i+l_i+M_{sl}\right) 
        \leq \bm{u}_i\left(s_{j}-\frac{\Delta t}{2}.v_{j}\right) 
\end{equation}
for all vehicles $i$ in sequence $q^*$ and for all vehicles $j$ in a conflicting group to vehicle $i$.

Finally, we enforce reachability by bounding the command velocities by the speed limit, $\bar v$, as well as the acceleration and breaking capabilities of each individual vehicle.

\begin{equation}
\label{eq:eq_cons_1}
    0\leq \bm{u}_i(t)\leq \bar v
\end{equation}
\begin{equation}
\label{eq:eq_cons_2}
    a^{min}_i \Delta t\leq \bm{u}_i(t)-v_i(t)\leq a^{max}_i \Delta t
\end{equation}

\noindent\textbf{Solving \optname:} \textit{The constraints (\ref{eq: long_linear}), (\ref{eq: lateral_linear}), (\ref{eq:eq_cons_1}), and (\ref{eq:eq_cons_2}) are provided to \optname with objective function given by Equation~(\ref{eq:J}), which is solved numerically using Gurobi (version 9.1.2) \cite{gurobi}.}

The computed target $\bm{u}_i$ command velocities are transmitted to each of the CAVs in the control zone. Note that we do not consider delays in communication and assume the data is transmitted instantaneously via V2I communication. The low-level controller on-board each CAV then computes the optimal acceleration or deceleration needed, to achieve the desired $\bm{u}_i$ command velocity within the control time ($\Delta t$) duration. 

Finally, a key feature of our approach is that it is robust to drift in the executed control. Our approach operates at a cycle frequency of $100$ ms after which new command velocities are obtained, updating the previous controls. Therefore, any deviation between optimal control and the executed control is erased during every $100$ ms update cycle.

\vspace{-5pt}
\section{Experiments and Results}
\label{sec: experiments_and_results}
\begin{figure*}[t]
\centering
   \begin{subfigure}[h]{0.325\textwidth}
    \includegraphics[width=\textwidth]{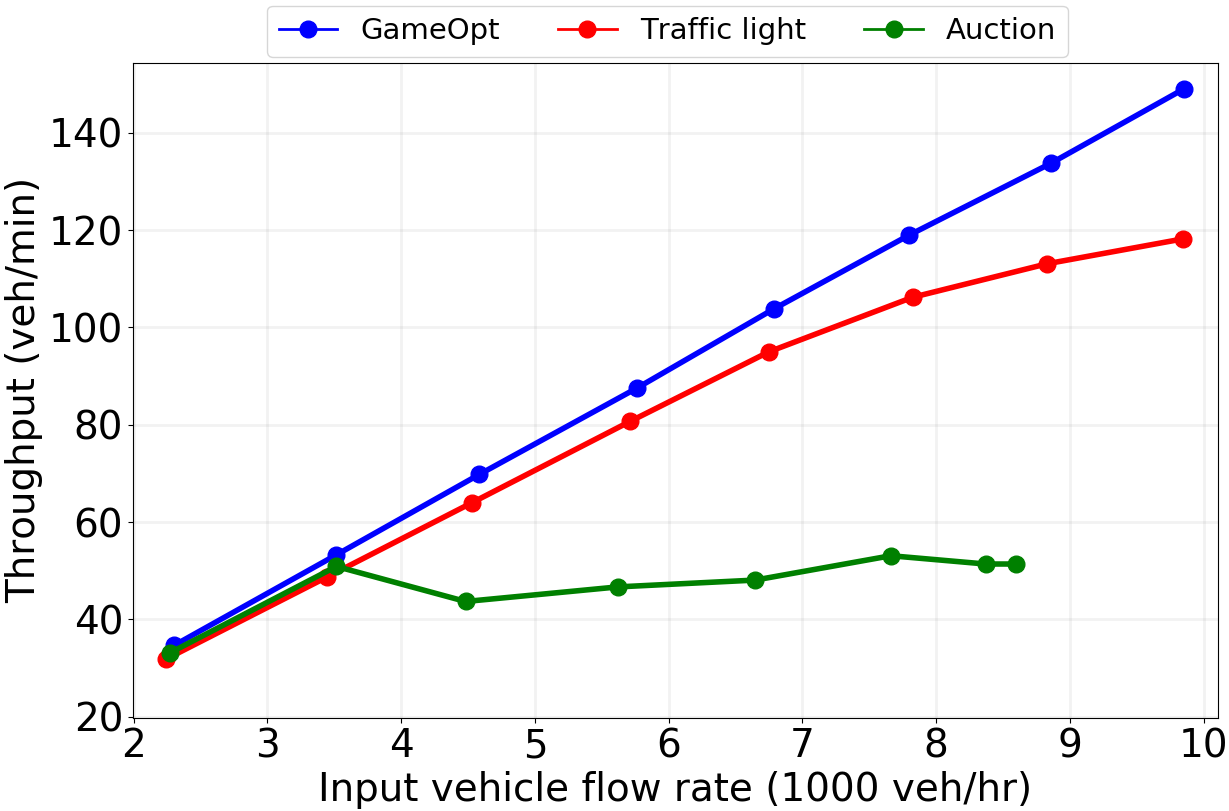}
    \caption{Throughput}
    \label{fig: Throughput}
  \end{subfigure}
 \begin{subfigure}[h]{0.325\textwidth}
    \includegraphics[width=\textwidth]{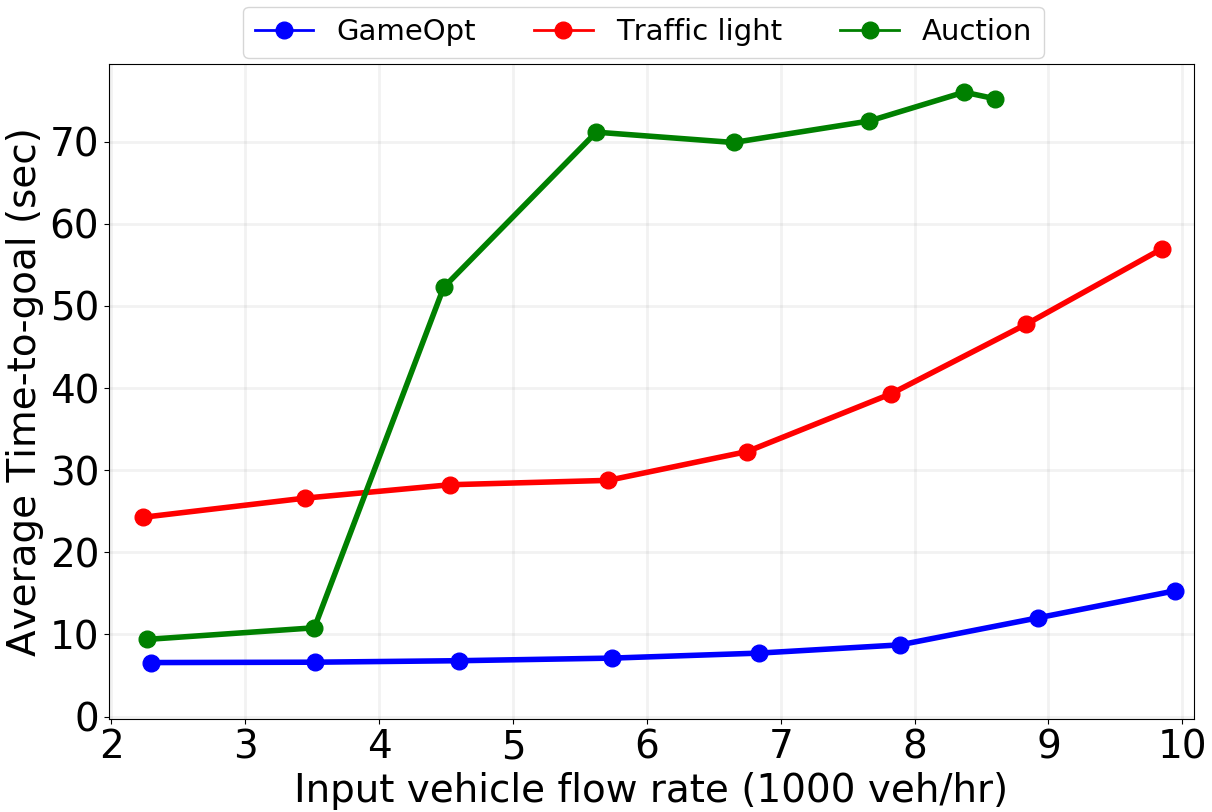}
    \caption{Time-to-goal}
    \label{fig: ttg}
  \end{subfigure}
 \begin{subfigure}[h]{0.325\textwidth}
    \includegraphics[width=\textwidth]{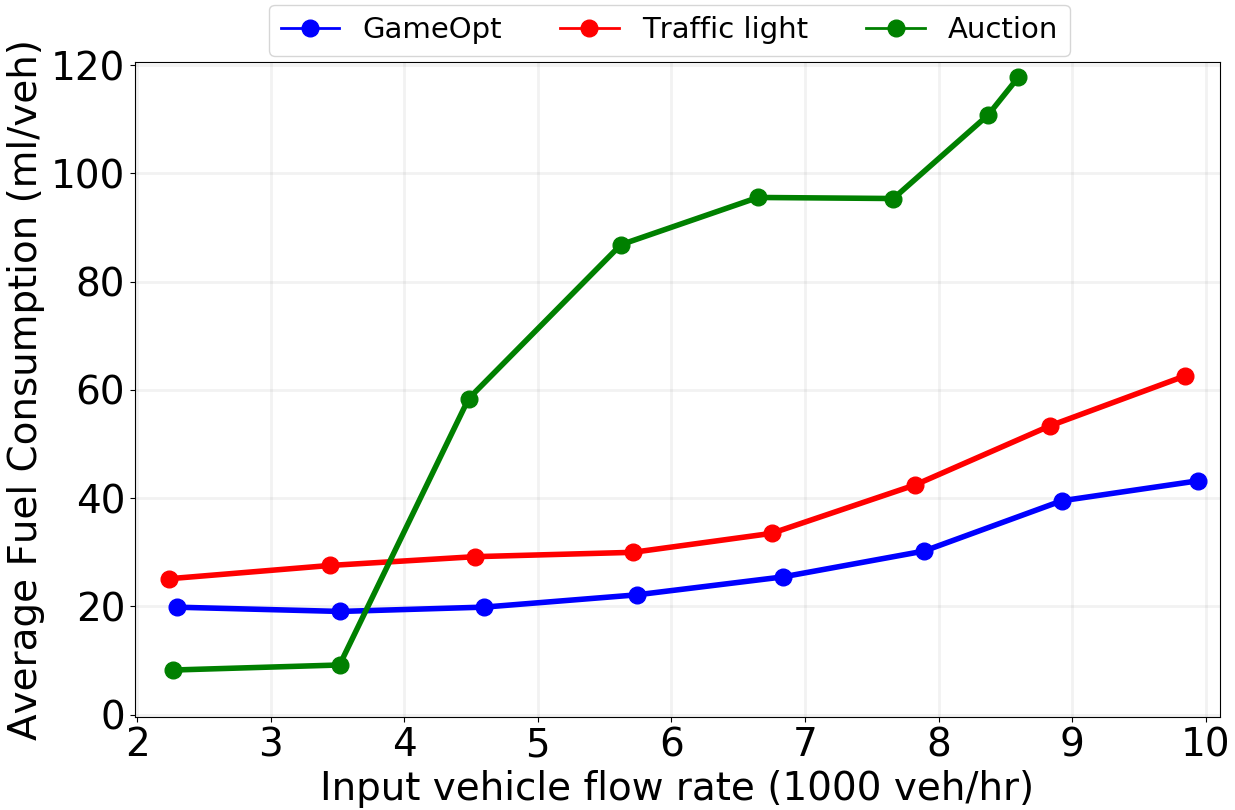}
    \caption{Fuel Efficiency}
    \label{fig: fuel_efficiency}
  \end{subfigure}

\caption{\textbf{Comparison with baselines:} \model increases throughput by approximately $8\%$ and $60\%$ compared to \textsc{Traffic-Light} and \textsc{Auction}, respectively. Furthermore, our method reduces the time-to-goal by $75\%$ and $80\%$, respectively, and improves fuel efficiency by $325\%$ and $66\%$, respectively, compared with \textsc{Traffic-Light} and \textsc{Auction}.
}
  \label{fig: results}
  \vspace{-15pt}
\end{figure*}

In this section, we evaluate the capabilities of our approach, study the impact of varying environmental parameters, and perform comparisons against other real-time capable methods.
\vspace{-15pt}
\subsection{Experiment Setup}
\label{subsec: experiment_setup}

We evaluate the performance of \model in a multi-lane four-arm intersection simulation using the SUMO platform~\cite{SUMO2018} with the following parameters; \textit{Control zone length ($L_c$) = $150$m}, \textit{Speed limit ($\bar v$) = $20$m/s} \textit{Objective trade-off ($\lambda$) = $0.7$}, \textit{Safety margins: $M_{sr}$ = $2.0$ and $M_{sl}$ = $25.0$}. The high-level controller uses the TraCI traffic controller interface to communicate with the simulation. All simulations and optimization algorithms run on a personal computer with an Intel i7-8750H CPU and 32GB of RAM. We provide results in additional scenarios with varying lanes, traffic density, and number of vehicles in~\cite{gameoptPage}.


\vspace{-5pt}
\subsection{Evaluation Metrics and Baselines}

The key performance metrics used to evaluate merging algorithms are \emph{throughput} (number of vehicles that can travel through the intersection per min), \emph{time-to-goal} (average time vehicles spend in the control zone) and \emph{fuel consumption}. 
To showcase the properties and appropriately assess the performance
of \model, we compare it with two other real-time capable baseline approaches.  
The first baseline we choose to compare with is an auction framework where the bidding strategy is based on the order in which agents arrive at the intersection, similar to the FIFO principle. To test the limits of our approach, we also compare against well-tuned traffic lights (light timing set to maximize performance), a well-established signaled intersection traffic management system. We refer to these baselines as \textsc{Traffic-Light} and \textsc{Auction} in the following sections.

\vspace{-5pt}

\subsection{Efficiency Analysis}

From the blue curves in Figure \ref{fig: results}, we observe that \model exhibits linearly increasing throughput with increasing input flow while the other two methods become capped after a point. Similarly, we see that \model is capable of maintaining low temporal delays as well as low fuel consumption across all values of input flow. This highlights the consistent performance capabilities of the \model method, even at very high input flow rates such as $10,000$ vehicles per hour. Note that input flow rate is defined as the total number of vehicles that enter into the intersection simulation per hour on all four roads.

\noindent\textbf{Results and Comparison with Baselines:} 
In Figure \ref{fig: results}, we show results of comparing \model with \textsc{Traffic-Light} and \textsc{Auction} based on throughput, time-to-goal, and fuel efficiency. Perhaps the most significant result is that not only does \model outperform the \textsc{Auction} approach, it also outperforms \textsc{Traffic-Light}, a well-established method optimized for signalized intersections. In Figure~\ref{fig: Throughput}, we observe that \model linearly increases in throughput (equivalent to maximizing flow through the intersection) with increasing flow rate. At high flow rates, \model performs approximately $25\%$ and $174\%$ better when compared to \textsc{Traffic-Light} and \textsc{Auction}, respectively. We also note that these relative performance improvements become more apparent with the increase in flow rate.  

This follows our intuition that with fewer vehicles, we do not require a complex optimization-based method; a simpler signal-based method suffices. However, as the traffic on the road increases, which often leads to increased congestion and delays, the benefit of \model becomes apparent. Furthermore, from Figures \ref{fig: ttg} and \ref{fig: fuel_efficiency}, we observe that at high flow rates our method reduces the time-to-goal by $75\%$ and $81\%$, respectively, and reduces fuel consumption by $33\%$ and $66\%$, respectively, compared with \textsc{Traffic-Light} and \textsc{Auction}. This shows that \model results in less waiting in queues along with less acceleration and braking tasks, resulting in improved fuel efficiency.

\begin{figure}[h!]
\vspace{-5pt}
\centering
   \begin{subfigure}[h]{.7\columnwidth}
    \includegraphics[width=\textwidth]{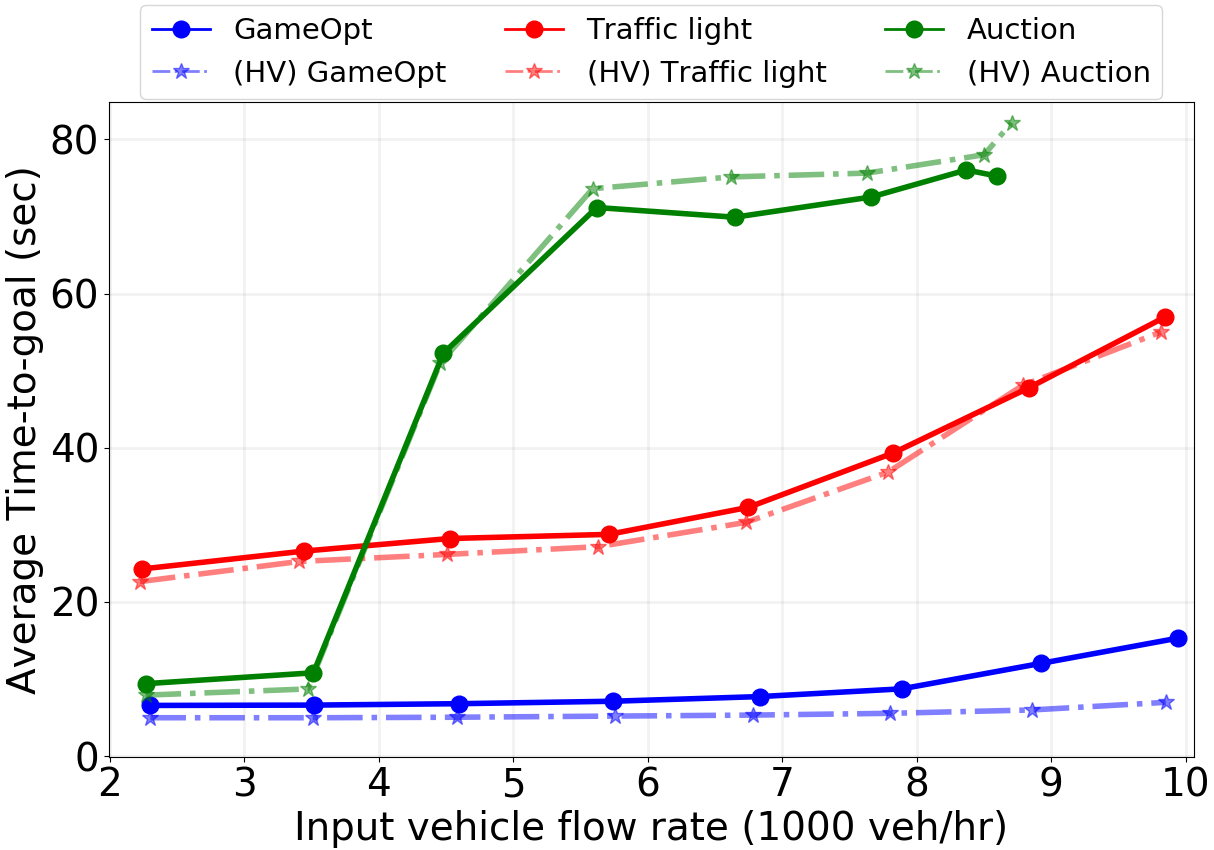}
    \caption{Impact of increased speed limit.}
    \label{fig: high_vel}
  \end{subfigure}
 \begin{subfigure}[h]{.7\columnwidth}
    \includegraphics[width=\textwidth]{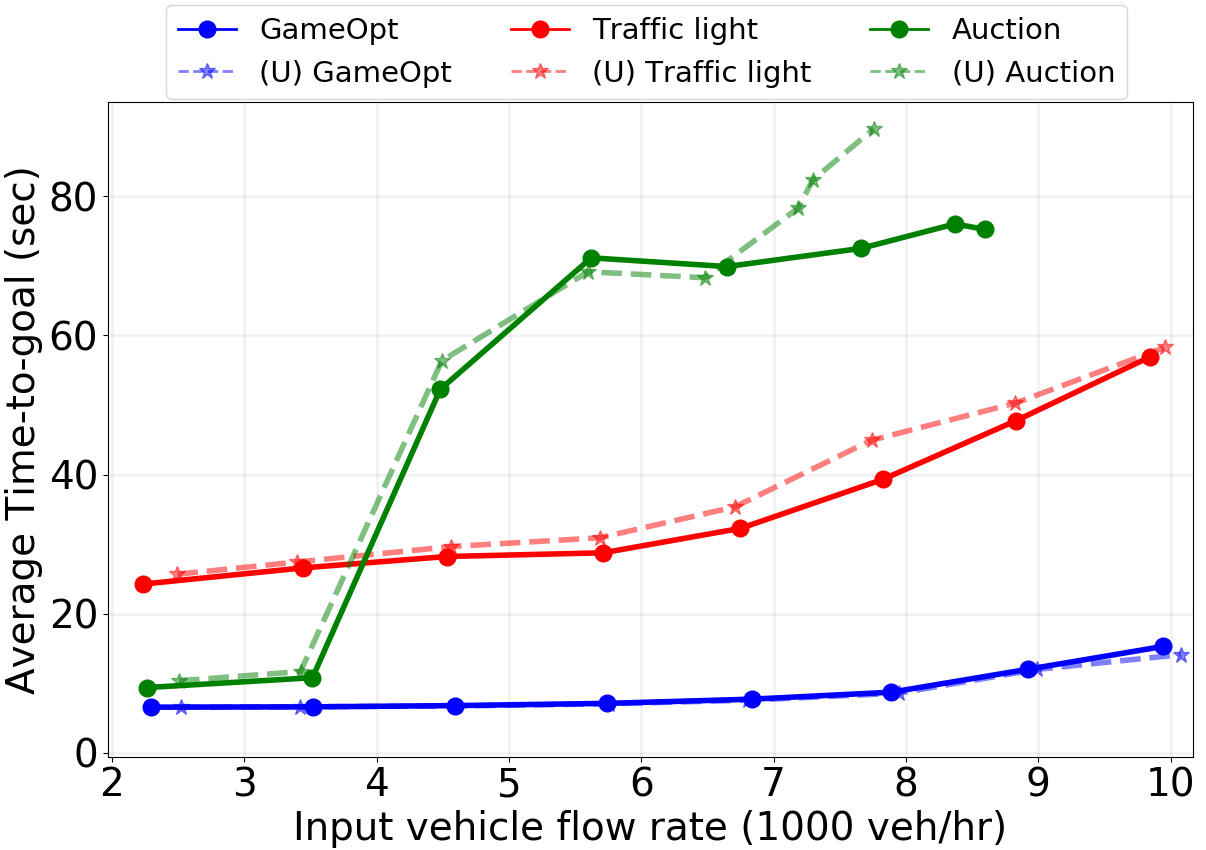}
    \caption{Impact of unbalanced inflow in different lanes.}
    \label{fig: unbalanced}
  \end{subfigure}

\caption{Impact of increasing the speed limit and unbalanced inflow rates on the average time-to-goal. \model handles both cases well and even performs better with a higher speed limit. \textsc{Traffic-Light} remains unchanged while \textsc{Auction} becomes worse in both cases.}
  \label{fig: speed_inflow_analysis} 
  \vspace{-5pt}
\end{figure}    

We also analyze the impact of increasing road speed limits from 20m/s to 25m/s, as this is a possibility in fully autonomous systems. In Figure~\ref{fig: high_vel}, we observe that while \textsc{Traffic-Light} and \textsc{Auction} both remain unchanged, \model exhibits reduced time-to-goal. This shows that the overall capabilities of \model can be increased further with improved road infrastructure. We also test the impact of creating unbalanced vehicle demand on different arms of the intersection by varying the inflow rate on each arm. Assuming we keep the total number of vehicles constant, some arms experience unusually high traffic density. As expected, from Figure~\ref{fig: unbalanced} we observe that \textsc{Auction}'s and \textsc{Traffic-Light}'s performance worsens whereas \model remains unchanged, demonstrating robustness to this configuration.

\noindent \textbf{Real-time Operation:} The algorithm in our optimization-based approach operates at a frequency of $f_s = 10$Hz (updates all control commands every $100$ms). The time required for the optimizer to compute control outputs for a single sequence has a mean of $1.16$ms, standard deviation of $1.73$ms, and a worst-case maximum of $17.25$ms. This is significantly lower than the controller update rate of $100$ms, thus enabling real-time operation. The fastest optimization-based method in literature takes up to $2.4$ seconds for $50$ vehicles~\cite{Seyed2018opt, nor2020optimal} in a simple single lane environment. In comparison, \model takes less than $18$ milliseconds for $50+$ vehicles in a multi-lane setting. This is a speed-up of around $130\times$ in considerably more complex environments.

\subsection{Effect of Overbidding and Underbidding}
\label{sec: overbid}
According to Theorem~\ref{thm: incentive_compatibility} in Section~\ref{subsec: optimality}, overbidding or underbidding yields a lower utility \cite{chandra2022gameplan}. In simulation, when the auction is combined with a trajectory planner, we observe even more drastic consequences. In the case of overbidding, the agent submits a higher than expected bid to the trajectory planner. The planner accommodates this higher bid by reducing a conflicting agent’s speed, causing the latter to slow down and create a dangerous situation with possible rear-end collisions. In the case of underbidding, the trajectory planner reduces the speed of the agent that is underbidding, resulting in possible rear-end collisions.

\section{Conclusion, Limitations, and Future Work}
\label{sec: conclusion}

We propose \model: a novel hybrid approach to cooperative intersection control for dynamic, multi-lane, unsignalized intersections. Our algorithm couples an auction mechanism that generates a priority entrance sequence with an optimization-based trajectory planner that computes the optimal velocity commands to achieve this sequence. This hybrid approach allowed us to achieve real-time computation capabilities in high-density multi-lane traffic, while providing guarantees in terms of efficiency, safety, and fairness. 
Our performance was verified using the SUMO platform, and we show that \model improves throughput by around $25\%$, time spent by $75\%$, and fuel consumption by $33\%$, compared to auction-based approaches and signaled approaches using traffic-lights and stop signs. We also demonstrate that \model operates at real-time speeds ($<10$ milliseconds), which is at least $100\times$ faster than prior methods.

Future work in this area would include exploring different kinds of auctions for flexibility in intersection control management, the impact of imperfect communication on the performance of the algorithm, and the capability to handle the needs of heterogeneous vehicles and mixed traffic.

{\footnotesize \bibliography{refs}}

\begin{thebibliography}{10}
\providecommand{\url}[1]{#1}
\csname url@samestyle\endcsname
\providecommand{\newblock}{\relax}
\providecommand{\bibinfo}[2]{#2}
\providecommand{\BIBentrySTDinterwordspacing}{\spaceskip=0pt\relax}
\providecommand{\BIBentryALTinterwordstretchfactor}{4}
\providecommand{\BIBentryALTinterwordspacing}{\spaceskip=\fontdimen2\font plus
\BIBentryALTinterwordstretchfactor\fontdimen3\font minus
  \fontdimen4\font\relax}
\providecommand{\BIBforeignlanguage}[2]{{%
\expandafter\ifx\csname l@#1\endcsname\relax
\typeout{** WARNING: IEEEtran.bst: No hyphenation pattern has been}%
\typeout{** loaded for the language `#1'. Using the pattern for}%
\typeout{** the default language instead.}%
\else
\language=\csname l@#1\endcsname
\fi
#2}}
\providecommand{\BIBdecl}{\relax}
\BIBdecl

\bibitem{grembek2018introducing}
O.~Grembek, A.~A. Kurzhanskiy, A.~Medury, P.~Varaiya, and M.~Yu, ``Introducing
  an intelligent intersection,'' \emph{Intelligent Transportation Systems
  Reports}, no.~13, 2018.

\bibitem{basic}
J.~Rios-Torres and A.~A. Malikopoulos, ``A survey on the coordination of
  connected and automated vehicles at intersections and merging at highway
  on-ramps,'' \emph{IEEE Transactions on Intelligent Transportation Systems},
  vol.~18, no.~5, pp. 1066--1077, 2017.

\bibitem{Nilesh2021merge}
N.~Suriyarachchi, F.~M. Tariq, C.~Mavridis, and J.~S. Baras, ``Real-time
  priority-based cooperative highway merging for heterogeneous autonomous
  traffic,'' in \emph{2021 IEEE International Transportation Systems Conference
  (ITSC)}, 2021, pp. 2019--2026.

\bibitem{Nilesh2021shockwave}
N.~Suriyarachchi and J.~S. Baras, ``Shock wave mitigation in multi-lane
  highways using vehicle-to-vehicle communication,'' in \emph{2021 IEEE 94th
  Vehicular Technology Conference (VTC2021-Fall)}, 2021, pp. 1--7.

\bibitem{chandra2022gameplan}
R.~Chandra and D.~Manocha, ``Gameplan: Game-theoretic multi-agent planning with
  human drivers at intersections, roundabouts, and merging,'' \emph{IEEE
  Robotics and Automation Letters}, 2022.

\bibitem{carlino2013auction}
D.~Carlino, S.~D. Boyles, and P.~Stone, ``Auction-based autonomous intersection
  management,'' in \emph{16th International IEEE Conference on Intelligent
  Transportation Systems (ITSC)}.\hskip 1em plus 0.5em minus 0.4em\relax IEEE,
  2013, pp. 529--534.

\bibitem{sayin2018information}
M.~O. Sayin, C.-W. Lin, S.~Shiraishi, J.~Shen, and T.~Ba{\c{s}}ar,
  ``Information-driven autonomous intersection control via incentive compatible
  mechanisms,'' \emph{IEEE Transactions on Intelligent Transportation Systems},
  vol.~20, no.~3, pp. 912--924, 2018.

\bibitem{rey2021online}
D.~Rey, M.~W. Levin, and V.~V. Dixit, ``Online incentive-compatible mechanisms
  for traffic intersection auctions,'' \emph{European Journal of Operational
  Research}, 2021.

\bibitem{gt6}
N.~Buckman, A.~Pierson, W.~Schwarting, S.~Karaman, and D.~L. Rus, ``Sharing is
  caring: Socially-compliant autonomous intersection negotiation,'' 2020.

\bibitem{schwarting2019social}
W.~Schwarting, A.~Pierson, J.~Alonso-Mora, S.~Karaman, and D.~Rus, ``Social
  behavior for autonomous vehicles,'' \emph{Proceedings of the National Academy
  of Sciences}, vol. 116, no.~50, pp. 24\,972--24\,978, 2019.

\bibitem{li2020game}
N.~Li, Y.~Yao, I.~Kolmanovsky, E.~Atkins, and A.~R. Girard, ``Game-theoretic
  modeling of multi-vehicle interactions at uncontrolled intersections,''
  \emph{IEEE Trans. on Intelligent Transportation Systems}, 2020.

\bibitem{tian2020game}
R.~Tian, N.~Li, I.~Kolmanovsky, Y.~Yildiz, and A.~R. Girard, ``Game-theoretic
  modeling of traffic in unsignalized intersection network for autonomous
  vehicle control verification and validation,'' \emph{IEEE Transactions on
  Intelligent Transportation Systems}, 2020.

\bibitem{Malik2019unconstrained}
A.~A. Malikopoulos and L.~Zhao, ``Optimal path planning for connected and
  automated vehicles at urban intersections,'' in \emph{IEEE Conference on
  Decision and Control (CDC)}, 2019, pp. 1261--1266.

\bibitem{Bian2020opt}
Y.~Bian, S.~E. Li, W.~Ren, J.~Wang, K.~Li, and H.~X. Liu, ``Cooperation of
  multiple connected vehicles at unsignalized intersections: Distributed
  observation, optimization, and control,'' \emph{IEEE Transactions on
  Industrial Electronics}, vol.~67, no.~12, pp. 10\,744--10\,754, 2020.

\bibitem{Pei2021spaceopt}
H.~Pei, Y.~Zhang, Y.~Zhang, and S.~Feng, ``Optimal cooperative driving at
  signal-free intersections with polynomial-time complexity,'' \emph{IEEE
  Transactions on Intelligent Transportation Systems}, pp. 1--13, 2021.

\bibitem{Seyed2018opt}
S.~A. Fayazi and A.~Vahidi, ``Mixed-integer linear programming for optimal
  scheduling of autonomous vehicle intersection crossing,'' \emph{IEEE
  Transactions on IV}, vol.~3, no.~3, pp. 287--299, 2018.

\bibitem{capasso2021end}
A.~P. Capasso, P.~Maramotti, A.~Dell'Eva, and A.~Broggi, ``End-to-end
  intersection handling using multi-agent deep reinforcement learning,''
  \emph{arXiv preprint arXiv:2104.13617}, 2021.

\bibitem{roh2020multimodal}
J.~Roh, C.~Mavrogiannis, R.~Madan, D.~Fox, and S.~S. Srinivasa, ``Multimodal
  trajectory prediction via topological invariance for navigation at
  uncontrolled intersections,'' \emph{arXiv preprint arXiv:2011.03894}, 2020.

\bibitem{roughgarden2016twenty}
T.~Roughgarden, \emph{Twenty lectures on algorithmic game theory}.\hskip 1em
  plus 0.5em minus 0.4em\relax Cambridge University Press, 2016.

\bibitem{wang2020game}
M.~Wang, N.~Mehr, A.~Gaidon, and M.~Schwager, ``Game-theoretic planning for
  risk-aware interactive agents,'' \emph{IROS}, 2020.

\bibitem{clrs}
T.~H. Cormen, C.~E. Leiserson, R.~L. Rivest, and C.~Stein, \emph{Introduction
  to algorithms}, 2009.

\bibitem{gameoptPage}
\BIBentryALTinterwordspacing
GameOpt, ``Gameopt project supplementary material,'' 2022. [Online]. Available:
  \url{https://gamma.umd.edu/gameopt}
\BIBentrySTDinterwordspacing

\bibitem{gurobi}
\BIBentryALTinterwordspacing
L.~Gurobi~Optimization, ``Gurobi optimizer reference manual,'' 2021. [Online].
  Available: \url{http://www.gurobi.com}
\BIBentrySTDinterwordspacing

\bibitem{SUMO2018}
P.~A. Lopez, M.~Behrisch, L.~Bieker-Walz, J.~Erdmann, Y.-P. Fl{\"o}tter{\"o}d,
  R.~Hilbrich, L.~L{\"u}cken, J.~Rummel, P.~Wagner, and E.~Wie{\ss}ner,
  ``Microscopic traffic simulation using sumo,'' in \emph{IEEE International
  Conference on Intelligent Transportation Systems}.\hskip 1em plus 0.5em minus
  0.4em\relax IEEE, 2018.

\bibitem{nor2020optimal}
M.~H.~M. Nor and T.~Namerikawa, ``Optimal motion planning of connected and
  automated vehicles at signal-free intersections with state and control
  constraints,'' \emph{SICE Journal of Control, Measurement, and System
  Integration}, vol.~13, no.~2, pp. 30--39, 2020.

\end{thebibliography}
\bibliographystyle{IEEEtran}

\end{document}